\documentclass{ecai} 

\usepackage{latexsym}
\usepackage{amssymb}
\usepackage{amsmath}
\usepackage{amsthm}
\usepackage{float} 
\usepackage{booktabs}
\usepackage{enumitem}
\usepackage{graphicx}
\usepackage{color}
\usepackage{algorithm}
\usepackage{algorithmic}
\usepackage{soul}
\usepackage{fancyhdr}
\usepackage{url}
\usepackage{colortbl}
\usepackage[utf8]{inputenc}
\usepackage{makecell}
\usepackage{adjustbox}
\usepackage{tcolorbox}
\usepackage{mathtools}
\usepackage{multirow}
\usepackage{multicol}
\usepackage{scontents}
\usepackage{lipsum}
\usepackage{tabularx}
\usepackage{cleveref}
\usepackage{array}
\usepackage{caption}
\newcolumntype{L}[1]{>{\raggedright\arraybackslash}p{#1}}

\theoremstyle{definition}
\newtheorem{definition}{Definition}
\usepackage{listings} 
\usepackage[colorinlistoftodos, textwidth=4cm, loadshadowlibrary]{todonotes}

\newcommand{\RNum}[1]{\expandafter{\romannumeral #1\relax}}
\newcommand{\thickhline}{%
    \noalign {\ifnum 0=`}\fi \hrule height 1pt
    \futurelet \reserved@a \@xhline
}

\definecolor{aquamarine}{rgb}{0.0, 0.8, 0.6}

\begin{document}
\begin{frontmatter} 

\title{Enabling MCTS Explainability for Sequential Planning Through Computation Tree Logic} 

\author[A]{\fnms{Ziyan}~\snm{An}}
\author[B]{\fnms{Hendrik}~\snm{Baier}}
\author[A]{\fnms{Abhishek}~\snm{Dubey}} 
\author[A]{\fnms{Ayan}~\snm{Mukhopadhyay}} 
\author[A]{\fnms{Meiyi}~\snm{Ma}} 

\address[A]{Vanderbilt University}
\address[B]{Eindhoven University of Technology}

\paperid{471}

\begin{abstract}
    Monte Carlo tree search (MCTS) is one of the most capable online search algorithms for sequential planning tasks, with significant applications in areas such as resource allocation and transit planning. 
Despite its strong performance in real-world deployment, the inherent complexity of MCTS makes it challenging to understand for users without technical background. 
This paper considers the use of MCTS in transportation routing services, where the algorithm is integrated to develop optimized route plans. These plans are required to meet a range of constraints and requirements simultaneously, further complicating the task of explaining the algorithm's operation in real-world contexts. 
To address this critical research gap, we introduce a novel computation tree logic-based explainer for MCTS. Our framework begins by taking user-defined requirements and translating them into rigorous logic specifications through the use of language templates. Then, our explainer incorporates a logic verification and quantitative evaluation module that validates the states and actions traversed by the MCTS algorithm. The outcomes of this analysis are then rendered into human-readable descriptive text using a second set of language templates. 
The user satisfaction of our approach was assessed through a survey with 82 participants. The results indicated that our explanatory approach significantly outperforms other baselines in user preference. 
\end{abstract}
\end{frontmatter} 


\section{Introduction}

Artificial Intelligence (AI) is now intricately woven into the fabric of our daily lives. AI's influence extends into virtually every sector, redefining the way we work, learn, and interact with the world around us. For example, AI algorithms play a pivotal role in personalized recommendations, virtual social networks, medical diagnosis, and transportation management. As AI becomes ubiquitous, explainability of the underlying algorithmic processes is imperative for fostering trust and ensuring transparency in automated decision-making processes, thereby enabling stakeholders to comprehend and effectively scrutinize autonomous and data-driven systems. The importance of developing transparent and explainable AI systems has been highlighted and, to some extent, mandated in public policy---both the European Union and the US have highlighted its importance through legislation~\cite{execOrder, panigutti2023role}.

The field of Explainable AI has acknowledged this challenge and developed several novel and effective approaches. Indeed, there are several well-known approaches for interpreting learning-based methods, i.e., methods that use historical data to make estimates about a dependent variable as a function of independent variables, such as SHAP~\cite{lundberg2017unified} and LIME~\cite{ribeiro2016should}. Unfortunately, there is a major lack of approaches that can explain sequential decision-making. Such AI-based methods significantly differ from standard supervised learning since the algorithms for sequential decision-making conduct a search through a complex combinatorial space of trajectories (i.e., sequential states and actions) to find optimal policies; as a result, standard approaches that can be used to explain supervised learning are not directly applicable to sequential decision making. 

\begin{figure}[t]
\centering
  \includegraphics[width=0.8\linewidth]{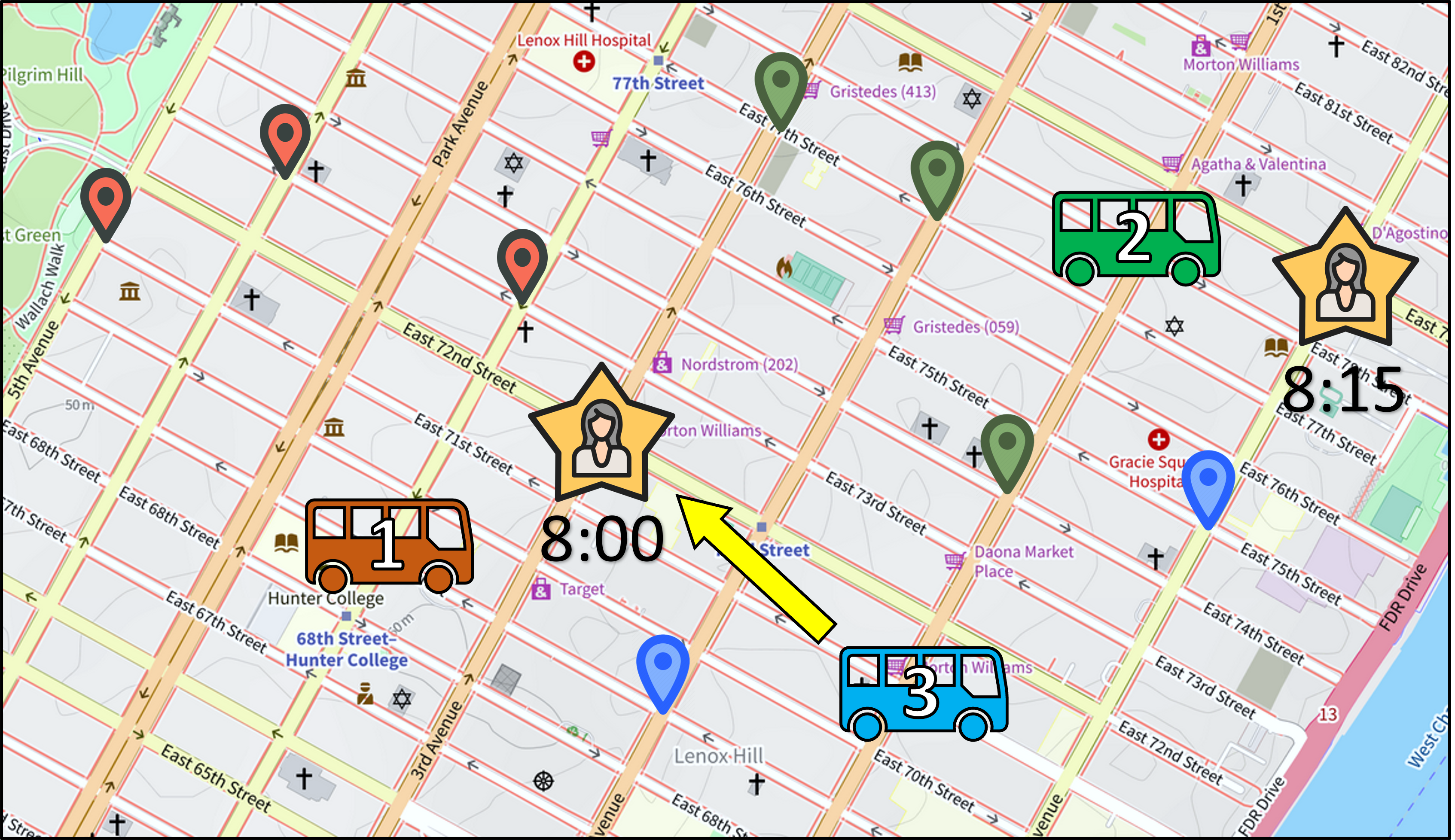}
  \caption{A running example of a trip request and the recommended route plan. Available vehicles are denoted by \textbf{bus icons} and the outstanding request is represented by the \textbf{passenger icon}. }
  \label{fig:runningeg}
  \vspace{3em}
\end{figure}

In this paper, we focus on online search, which is at the core of recent advances in autonomous planning and decision-making~\cite{schrittwieser2020mastering}. Specifically, we focus on Monte Carlo tree search (MCTS)~\cite{kocsis2006bandit}, an online and anytime sampling-based general search algorithm that has played a critical role in achieving state-of-the-art performance in games such as Chess and Go~\cite{silver2018general} and also in many real-world domains such as emergency response~\cite{mukhopadhyay2019online}, public transit~\cite{wilbur2022online}, and supply chain management~\cite{claes2017decentralised}. While \citeauthor{baier2021towards} highlight the need for explainable search~\cite{baier2020explainable, baier2021towards}, there are \textit{no well-established approaches} for explaining decisions computed by MCTS. 

To ground our framework, we use a complex real-world domain instead of board games.
In particular, we choose public transportation, where algorithmic and data-driven approaches have shown particularly promising results~\cite{shah2020neural,wilbur2022online}. However, transportation has traditionally involved human operators optimizing decisions manually; as a result, AI-based approaches are often viewed with skepticism even when they demonstrate higher efficiency. This lack of trust primarily stems from a lack of understanding of the algorithms. 

Consider a scenario in dynamic vehicle routing. As illustrated in Figure~\ref{fig:runningeg}, the path planning algorithm is responsible for assigning a passenger request to one of three available vehicles. The algorithm is equipped with real-time information, including the current locations of all vehicles and their availability status. While generating \textit{plans}, the algorithm is guided by specific objectives, such as maximizing the overall percentage of completed trips. Moreover, it is provided with certain non-negotiable constraints that must not be violated during the planning process. 
Leveraging its internal mechanisms, the planning algorithm generates the best approximation to an optimal solution it can find and presents the user with the corresponding result, which is to assign the passenger to the blue vehicle. 

In this case, the algorithm is a black-box when presented to users such as route dispatchers who may not have a technical background in search algorithms. 
Furthermore, the dispatchers remain uninformed about the potential future status of other vehicles in the vicinity. 
Without this knowledge, they are unable to determine if alternative nearby vehicles could provide better solutions to the current routing problem, which might also significantly benefit future operations. We tackle this challenge by developing computation tree logic (CTL)-based explainable MCTS. 
Particularly, our focus is on non-technical route dispatchers as explainees, who are seeking to understand the decision-making outcomes.
They may raise inquiries such as ``why was this particular action recommended?'' or ``why wouldn't this alternative action be viable?''

\begin{figure}[t]
\centering
  \includegraphics[width=\linewidth]{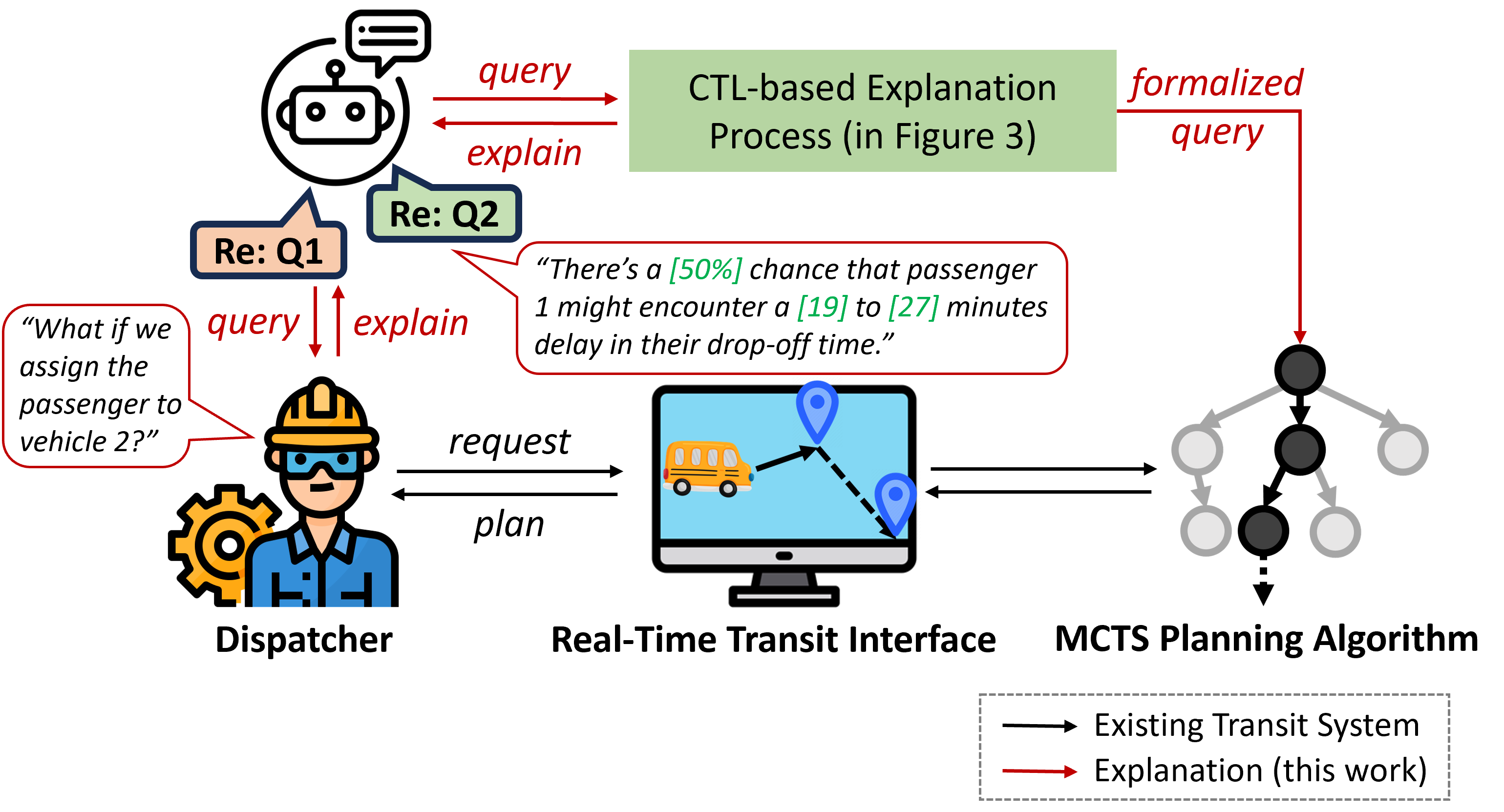}
  \caption{Overview of the CTL-based explainer.}
  \label{fig:overviewnew}
  \vspace{3em}
\end{figure}

Our explainer categorizes user queries into three varieties: \textit{factual queries} that provide insights into the algorithm's decisions derived from states and actions in the search tree, \textit{contrastive queries} that compare user-suggested decisions with the algorithm's recommendations by identifying CTL violations or inferior rewards, and \textit{alternative plan explanations} that involve extending the search to additional options. 
Specifically, our explainer first leverages language templates to capture user queries about the outcomes of the MCTS algorithm.
This design choice considers the needs of non-technical dispatchers by helping them in formulating the most relevant questions, excluding queries that do not directly relate to the core aspects of the planning problem. 
Second, each templated natural language query is converted into a logic specification by a many-to-many mapping process. 
For instance, consider a user query regarding time constraints in the proposed route plan from the previous example, such as, \textit{``Based on the suggested vehicle assignment, is it expected that the passenger will be dropped off too late?''} Our explainer translates this query into CTL as \textit{$\phi = AG \, ( t_{ \text{est.}} > t_{\text{allowed}})$}. 

Third, our explainer triggers a formal specification checking mechanism of CTL that takes logic formulas as input and traverses the MCTS tree to generate results. Fourth, the checking and verification results obtained in this step are then translated back into natural language. This conversion, again, leverages pre-defined language templates that are designed to ensure the output is understandable to the user. Continuing from the previous example, a response to the query would be: ``The passenger has specified a desired drop-off time of 5:33 PM. The route planning algorithm has simulated approximately 150 potential future scenarios and requests. \textit{Potential Late Arrival:} There's a chance that the passenger might encounter a delay in their drop-off time. This expected delay averages around 23 minutes. The primary reason for this delay is that the proposed vehicle is expected to make stops at about 4 other locations prior to reaching the passenger's drop-off point. However, the delay can be as short as 19 minutes or extend up to 27 minutes. The percentage of times the suggested vehicle doesn't meet the desired drop-off time is about 10\%.''
In summary, the CTL-based explainer offers the following \textit{contributions:} 
\begin{itemize}[left=0pt]
    \item Our explainer dynamically associates non-technical user queries with three types. These queries, containing free variables, are then converted into CTL formulas for further analysis. 
    \item Our explainer utilizes CTL semantics to determine the satisfiability of the logic query. It then translates these results into natural language using linguistic templates, ensuring the algorithm's explanations are easily comprehensible and accessible. 
    \item Our explainer was thoroughly evaluated in a study approved by the Institutional Review Board (IRB). We recruited a total of 82 participants for the study. The CTL-based explainer significantly outperforms the two baseline methods in reported understanding, satisfaction, completeness, and reliability.
\end{itemize}


\section{Sequential Planning in Public Transit}
In this section, we introduce the underlying Markov Decision Process (MDP) that models our specific use case in public transit, and defines the search space for our MCTS implementation.
Following related work~\cite{gupta2018planning,wilbur2022online} in using MDPs for transit planning, we define a transit planning task $\Pi$ as the tuple $\langle S, A, T, R, \gamma \rangle$, where $S$ is the set of states of the environment, $A$ the set of available actions for the planning agent, and $T(s_t, a_t, s_{t+1})$ defines the probabilities of moving from state $s_t$ to state $s_{t+1}$ under action $a_t$, where every transition to a next state implies a new predicted request. The reward function $R(s, a)$ and the discount factor $\gamma$ assign a numerical reward for taking action $a$ in state $s$.

A state $s$ in the transit planning task is defined by the tuple $ \langle \theta, r, V, \{ R^{\mathopen|V\mathclose|} \} \rangle $, where $\theta$ denotes the current route plans for all vehicles, $r$ is an outstanding request, $V$ denotes the locations of all vehicles, and $\{ R^{\mathopen|V\mathclose|} \}$ is a list of assigned requests to each vehicle~\cite{wilbur2022online}. 
We denote the state at time $t$ with $s_t = \langle \theta_t, r_t, V_t, \{ R^{\mathopen|V\mathclose|}_t \} \rangle$. 
An action $a_t$ at time $t$ involves assigning an outstanding trip request $r^j$ to a vehicle $v^i$. 
A transit service request $r^j$ is defined by: the time the request was made $t^j_r$, requested pickup time $t^j_p$, requested drop-off time $t^j_d$, pickup location $l^j_p$, and drop-off location $l^j_d$, as well as its current status $u^j \in \{\text{waiting},\, \text{assigned},\, \text{in-transit},\, \text{dropped-off}\}$. 
Each transit vehicle $v^i \in V$ has a fixed capacity $c^i$ and an occupancy $p^i$ that varies based on its current route plan. The designated route plan of each vehicle is denoted as $\theta^i$, specifying a list of future locations that the vehicle is scheduled to visit. 

\begin{table*}[t]
\centering
    \scriptsize
    \caption{Examples of queries, state variables, and the completed CTL formulas.}
    \begin{tabular}{L{2.5cm}|L{5.3cm}|L{4.7cm}|L{3cm}}
    \toprule
      \textbf{Query Type} & \textbf{Query Example} & \textbf{State Variable} & \textbf{CTL Formula} \\
      \midrule
      \textbf{T1:} Factual query & \textbf{Q1:} Is it expected that the [passenger] will be [picked up] on time? (Efficiency) & $t_{\text{est}}$ estimated travel time; $t_p$ request picked up time; $t_a$ allowed time window & $\phi_1: AG \, ( t_{\text{est}} \leq (t_p + t_a))$\newline  \\
    \midrule
    \textbf{T2:} Contrastive query & \textbf{Q2:} Why wasn't the passenger assigned to [this alternative vehicle]? (Efficiency \& Hard constraint) & $v_c$ vehicle capacity; $v_o$ vehicle occupancy; \newline $t_d$ request drop off time; $t_{\text{est}}$; $t_p$; $t_a$  & $\phi_1$; $\phi_2: AG \, (v_o \leq v_c )$;\newline $\phi_3: AG \, ( t_{\text{est}} \leq (t_d + t_a))$ \\
    \midrule
    \textbf{T3:} Query with tree expansion & \textbf{Q3:} Can you tell me more about [this alternative route]? (Efficiency \& Hard constraint \& Soundness) & ${v}_{\text{tt}}$ total travel time; ${v}_{\text{rt}}$ reasonable timeframe; \newline $t_{\text{est}}$; $t_p$; $t_d$; $t_a$; $v_c$; $v_o$ & $\phi_1$; $\phi_2$; $\phi_3$; $\phi_4: AG \, ({v}_{\text{tt}} \leq {v}_{\text{rt}})$ \\
      \bottomrule
    \end{tabular}
     \label{tab:overall_query_table}
\end{table*}

We use Monte Carlo Tree Search (MCTS) \cite{kocsis2006bandit} as the decision making algorithm, mapping the current state $s_t$ to the next action to take $a_t$ with the help of sampling-based planning over the MDP model described above. Our transit scenario operates within a dynamic environment where requests can be made at any moment. When a new request $r$ is initiated by a passenger, the planning algorithm is engaged on-the-fly to generate a new action to accommodate this request. This time point is referred to as a ``decision epoch''~\cite{joe2020deep,wilbur2022online}, where each decision epoch addresses a new planning problem independently. Specifically, MCTS seeks to find or approximate the optimal action $a_t$ that allocates the current request to one of the vehicles $v^i \in V$. 
Potential actions in this use case must adhere to two hard constraints: the capacity constraint $p^i_t \leq c^i, \; \forall v_i \in V$ and the timing constraint $\left| r^j_{\text{{dropped-off}}} - r^j_{\text{{in-transit}}} \right| \leq T_{\text{{max}}}, \; \forall r^j \in R$, where $p^i_t$ is the number of passengers for vehicle $v^i$ at time $t$ and $T_{max}$ is the maximum allowed en-route time for a trip request, $r^j_{\text{{dropped-off}}}$ is the actual drop off time and $r^j_{\text{{in-transit}}}$ is the actual pick up time of request $r^j$. 
A request is successfully assigned to a vehicle if the action complies with all hard constraints, which can extend beyond the above and are explicitly defined by the user or the algorithm engineer prior to the planning task. 
To simplify the problem, we do not consider swapping route plans between vehicles. Therefore, the action space of each vehicle is restricted to one per single request by inserting the pickup and drop-off locations of $r$ into its existing route at the index where the time used to travel between two stops is minimized. Assume that there are $\vert V \vert$ vehicles available, the maximum number of potential actions for assigning a particular request is $\vert V \vert$. 

The reward of the MCTS planning algorithm is below, where $N_t$ is the total number of in transit requests, $N_d$ is the number of dropped off requests, $\gamma_1$, $\gamma_2$, and $\gamma_3$ are weights. 
\begin{equation}\label{eq:obj}
\begin{aligned}
    &R(s,a) = \gamma_1 \cdot \frac{N_t + N_d}{\left| R^{\left| V \right|} \right|} +\gamma_2 \cdot \sum_{i=1}^{N_t + N_d} (t^i_p - r^i_{\text{{in-transit}}})\\
    & \quad\quad\quad\;  + \gamma_3 \cdot \sum_{i=1}^{N_d} (t^i_d - r^i_{\text{{dropped-off}}}) 
\end{aligned}
\end{equation}

\begin{figure}[t]
\centering
  \includegraphics[width=\linewidth]{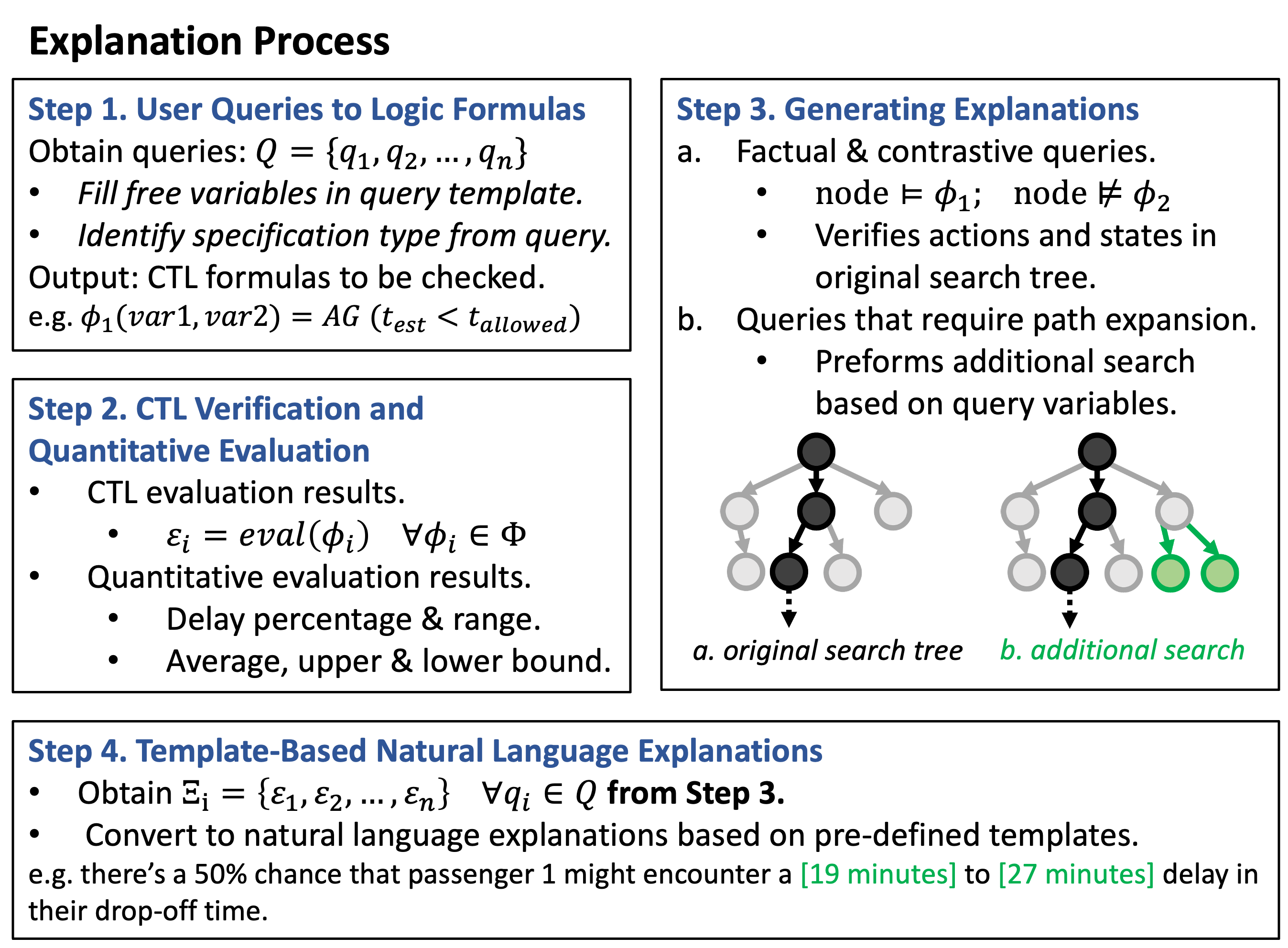}
  \caption{Illustration of the complete explanation process.}
  \label{fig:processoverview}
  \vspace{3em}
\end{figure}

\section{CTL-Based Explainable MCTS}
The goal of CTL~\cite{clarke1981design}-based explainable MCTS is to help users (e.g., transit operators) understand the relationship between the system state, the desired outcome, non-negotiable constraints, and the route plan produced by MCTS. 
In each decision epoch of the planning task $\Pi$, the user is presented with the current system state $s_t$ and the action $a_t$ proposed by MCTS. The user also has knowledge of a predefined set of hard constraints that the recommended actions must consistently adhere to. 
Given this information, the user poses a set of queries, denoted as $Q = \{q_1, q_2, \cdots \}$. Following the running example in Figure~\ref{fig:runningeg}, one such query can be: ``Why wasn't the passenger assigned to the green vehicle's route?'' (Q2, Table~\ref{tab:overall_query_table}). 

Given the query, the explainer's goal is to provide explanations, denoted as $\Xi$, concerning recommended actions $a_t$ in response to specific queries.
The explainer must deduce answers to these questions based on criteria that include the goal of the search, the stringent state requirements such as vehicle capacity violations, and user-proposed criteria like additional efficiency requirements. In Figure~\ref{fig:processoverview}, we provide a overview of this process. In the following sections, we describe the detailed methods employed to derive this response from the MCTS search tree. 


\subsection{Formulating User Queries}
Dispatchers lacking technical knowledge in MCTS can inquire about various aspects of the planning result, such as constraint compliance, route plan efficiency, and its soundness. To address these varying interests, we organize user queries into three distinct categories. Table~\ref{tab:overall_query_table} demonstrates one example query of each category. In principle, our explanation framework can handle any query that can be structured in this way. In our examples, free variables are indicated by brackets. Furthermore, these queries are initially left empty and are associated with predefined criteria such as efficiency and soundness. When a user submits a query in natural language, the first step involves identifying the completed variables within the query. For instance, if a query asks about an alternative action, we map the free variables in the query to the corresponding state variables of the alternative action in question. Additionally, we incorporate these state variables and the query criteria into automatically generated CTL formulas, denoted as $\Phi = \{ \phi_1, \phi_2, \cdots \}$. In Section~\ref{sec:Handling-Queries-by-Type}, we describe how variables in queries are processed, incorporated into CTL formulas, and how their handling differs within the search tree. 

\subsection{CTL Verification and Quantitative Evaluation} 
Before discussing how specific queries and criteria are translated into CTL, we detail our approach to employing the branching-time logic extensively applied in model checking~\cite{bouyer2017timed}. The CTL state quantifiers evaluate the breadth of the computation tree, where the $A$ (for all) operator examines every child of a given state, and the $E$ (exists) operator focuses on at least one child, requiring the formula's validity for any one of them. In contrast, the path quantifiers consider the tree's depth. The $F$ (future) operator is concerned with a condition's eventual occurrence and the $G$ (globally) operator requires that a condition holds true in all future states. For a CTL formula to be well-formed, a state quantifier must be followed by a path quantifier, as defined in Definition~\ref{def:1}. 
\begin{definition}[Syntax of CTL formulas]
\begin{align}
& \Phi 
::= 
\top 
\mid \bot
\mid p
\mid \neg \Phi
\mid \Phi \wedge \Phi
\mid \Phi \vee \Phi \nonumber \mid \\ & 
 AX\Phi
\mid EX\Phi 
\mid AF\Phi
\mid EF\Phi 
\mid AG\Phi
\mid EG\Phi   \nonumber
\end{align}\label{def:1}
\vspace{-1.5em}
\end{definition}

Given a CTL formula and a MCTS search tree, the process of verifying whether the nodes and paths satisfy the formula is conducted through CTL semantics, known as CTL verification or specification checking. 
The input for this specification checking module is a finite sequence of possible states of the system $\{ s_i \}, \text{ where } i < L$ and $L$ represents the length of the longest path within the data structure. This sequence describes discrete transitions $s_i \curvearrowright s_{i+1}$ with a time delay between two successive states, such that $t_{i+1} = t_i + d$~\cite{alur1993model}. In our planning task, the search tree is conceptualized as a Kripke structure~\cite{kripke1963semantical} where nodes symbolize states $s_i$ in the decision-making process, and edges represent the available choices or actions at each node, leading to the next state $s_{i+1}$ in the sequence. 
The CTL specification checking module yields a binary outcome, indicating whether the evaluated condition is satisfied or violated, respectively. 
To provide a quantitative assessment of each state where the CTL formula evaluates to false, we record the specific details of the violation, such as the extent of delay in minutes, at that particular state. This state is then marked with a flag to indicate a violation. This approach facilitates a thorough identification of non-compliant states, both qualitatively and quantitatively. It also enables for a more detailed transition to natural language explanations in subsequent processes, ensuring a comprehensive understanding of each violation. 

\begin{figure}[t]
\centering
  \includegraphics[width=0.98\linewidth]{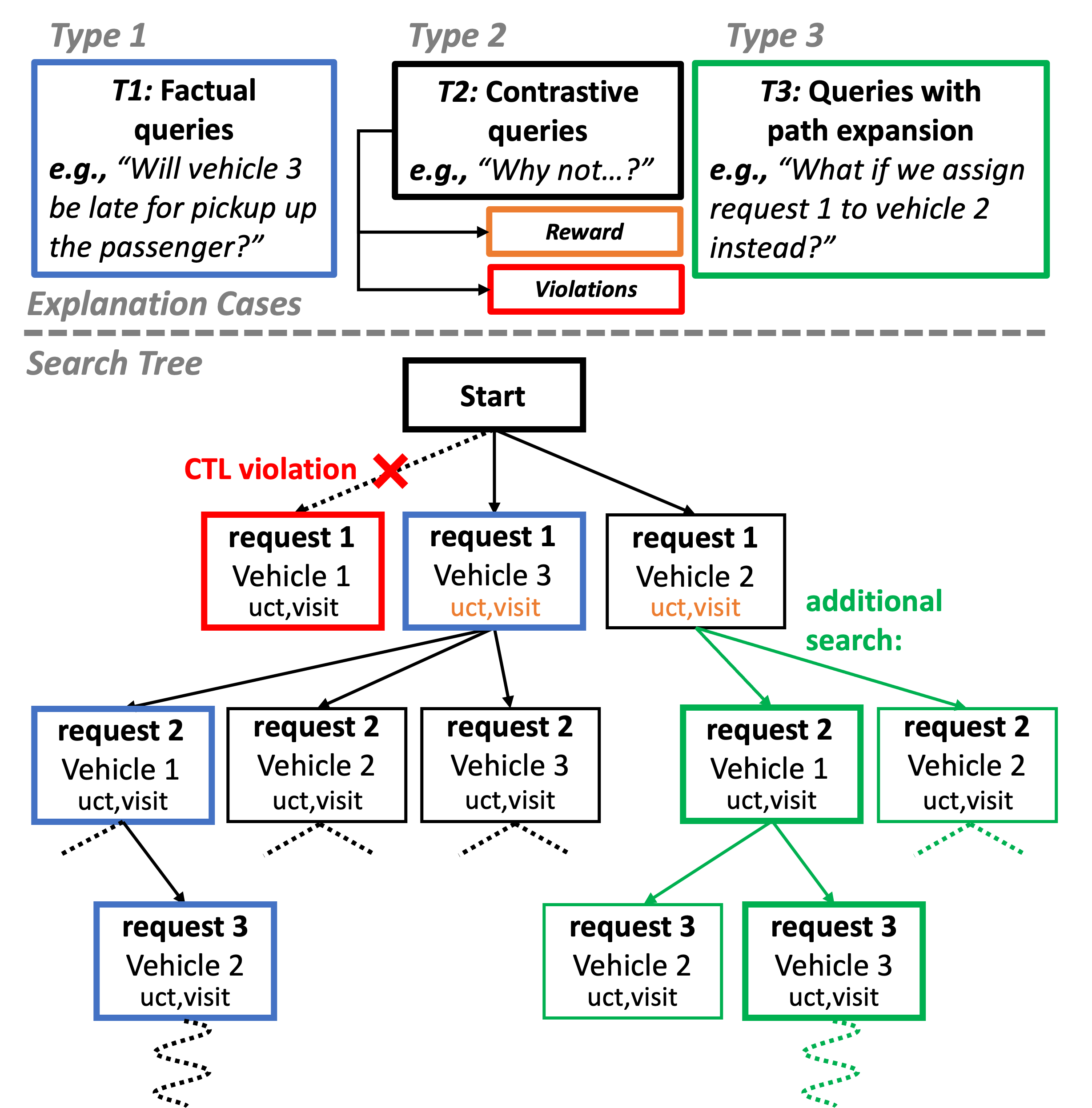}
  \caption{Illustration of different types of queries.}
  \label{fig:treevis}
  \vspace{3em}
\end{figure}

\subsection{Mapping Queries to CTL Formulas}\label{sec:Handling-Queries-by-Type}

\paragraph{Factual Queries --- ``Why?''}
Factual queries relate to reasons that indicate why a particular action is recommended~\cite{darwiche2022computation,shih2018symbolic}. 
We specifically address the following types of questions in which individuals seek to determine whether the specific user-specified criteria will be met. For instance, a common question is, ``Based on the current vehicle assignment, is it expected that passenger 1 will be dropped off too late?'' 
For such queries, we employ the following structured language template: ``Based on the current vehicle assignment, is it expected that [\textit{passenger number}] will be [\textit{action}] [\textit{time}]?'', where the \textit{action} can be either dropped off or picked up, and \textit{time} corresponds to either late or early. 
For the example query, we decompose the template-based query into the following tuple: [passenger 1, drop off, late]. 
As shown in Table~\ref{tab:overall_query_table}, the specification type for this factual query is ``Efficiency'', and the queried state variable is ``drop off'' or $t_d$. Thus, the corresponding CTL formula regarding efficiency would be $\phi_1: AG \, ( t_{\text{est}} \leq (t_d + t_{\text{allowed}}))$, where the variable $t_{\text{est}}$ represents the estimated time required for dropping off passenger 1. The CTL operator $AG$ checks for all states and all future paths starting from the root. The relation $\leq$ in the formula is used to determine whether passenger 1 will arrive late, based on the comparison of the estimated travel time with the allowed time window. 

\paragraph{Contrastive Queries --- ``Why Not?'' }
The second type of query in our explainer involves alternative plans specified by technical users. In this case, we aim to explain why the planning algorithm did not recommend the alternative action $a_t'$ as proposed by the user. We formulate the type of user queries about alternative plans with the following language template: ``Why wasn't [\textit{passenger}] assigned to [\textit{another vehicle}] located at [\textit{location}]?'', where the missing variables can be represented as [passenger, another vehicle, location]. For example, a common query is ``Why wasn't passenger 1 assigned to the red vehicle (which is closer)?'' 
In line with the nature of contrastive queries, the evaluation criteria for this query involve asking about potential violations of hard constraints and efficiency. More precisely, when querying about an alternative action $a_t'$, we can provide two types of reasons why MCTS did not produce $a_t'$. First, $a_t'$ could be infeasible due to violations of hard constraints. Second, $a_t'$ could be less efficient than the near-optimal solution $a_t$ generated by MCTS.
This evaluation includes several CTL formulas: $\phi_1: AG , ( t_{\text{est}} \leq (t_p + t_a))$, which examines efficiency in pickup times; $\phi_2: AG , ( t_{\text{est}} \leq (t_d + t_a))$, which addresses efficiency in drop-off times; and $\phi_3: AG , (v_o \leq v_c )$, which checks for violations of capacity constraints. Initially, the search tree nodes are checked against these hard constraints. If no violations are found, the efficiency formulas are then evaluated.

\paragraph{Queries with Tree Expansion --- ``What If?''}
MCTS focuses on exploring the most promising paths, intentionally leading to less exploration of other, lower-scoring branches.
Therefore, while extracting information from the search tree suffices for explaining recommended plans, it can provide limited insights for alternative plans, which might not have been explored enough \cite{baier2020explainable}. 
For example, consider a user querying about an alternative action, such as assigning request 1 to vehicle 2 in Figure~\ref{fig:runningeg}. In this case, while there are no direct violations, the user might question whether the algorithm considers the action less optimal due to insufficient exploration depth. 
To accommodate such queries, the MCTS explainer enables users to query about a broader set of potential actions that were neither recommended nor sufficiently explored.
We formulate them as ``Can you tell me more about assigning the [\textit{passenger}] to [\textit{another vehicle}] ?''. For instance, a user may ask, ``Can you tell me more about assigning passenger 1 to the green vehicle?'' In this format, the free variables are encapsulated in the tuple: [passenger 1, green vehicle]. This query type requires additional search and exploration to provide comprehensive explanations~\cite{baier2021towards}, as illustrated in Figure~\ref{fig:treevis}. 
A pseudocode of this process is provided in Appendix A1. 
At a high level, to address such queries, we locate the under-explored node $N(s,a)$ in the original tree, where $a$ is the alternative action being queried and $s$ is the state in the query. We then create a sub-tree by designating the previous path as the parent node and execute MCTS, continuing to explore possible scenarios and outcomes in response to the query.

\begin{algorithm}[t] 
\small
    \caption{\small CTL-based Explainable MCTS}
    \label{alg:process}
    \begin{flushleft}
    \begin{algorithmic}[1] 
    \FOR{each decision epoch}
        \STATE $s_t$ = $\langle \theta_t, r_t, V_t, \{ R_t^{\left|V\right|}\} \rangle$
        \\ \texttt{/* calculate route plan with MCTS */} 
        \STATE $a_t$ = MCTS($s_t$)
        \STATE Obtain user queries $Q_t =\{q_1, q_2, \cdots\}$
        \\ \texttt{/* process user queries */} 
        \FOR{$q_i \in Q_t$} 
        \STATE $\Phi \leftarrow \{ \}, \Xi \leftarrow \{ \}$
        \STATE Identify specification type from query type
        \STATE $\Phi \leftarrow \text{specification\_type}(q_i), \text{query\_variable}(q_i)$
        \STATE Calculate results: $\varepsilon_i = \textit{ExpGen}(\phi_i); \forall \phi_i \in \Phi$
        \STATE Append each $\varepsilon_i$ to $\Xi$; output $\Xi$
        \ENDFOR 
        \STATE Apply $a_t$ to route $\theta$
    \ENDFOR
    \end{algorithmic}
    \end{flushleft}
\end{algorithm}

\subsection{Generating Natural Language Explanations}
As outlined in Algorithm~\ref{alg:process}, the explainer is invoked once at the end of each decision epoch if there are user queries. Following the steps specified in Algorithm~\ref{alg:expgen}, the CTL-based explainer checks the search tree against each formula upon formulating the set of CTL formulas $\Phi$. Each explanation in $\Xi$ is obtained through CTL specification checking and quantitative evaluation. 
Specifically, at line 13 of Algorithm~\ref{alg:expgen}, we derive two key quantitative insights from the aggregated list of violations across all expanded nodes within the MCTS tree. First, we compute the violation percentage and the average degree of these violations. Second, for timing-related violations, we compute the temporal range of these violations, identifying both the earliest (lower bound) and latest (upper bound) occurrences.
These results are then translated into a set of natural language explanations, denoted as $\Xi = \{ \varepsilon_1, \varepsilon_2, \cdots \}$. 
The details of language templates are provided in Appendix A2.6. An example of these templates to address the query in the running example is: ``Based on the extensive set of scenarios examined by MCTS, there's a chance that the passenger might encounter a delay in their drop-off time. This anticipated delay averages around [23 minutes]. The primary reason for this delay is that the proposed vehicle is expected to make stops at about [4 other locations] prior to reaching the passenger's drop-off point. However, the delay can be as short as [19 minutes] or extend up to [27 minutes]. The percentage of times the suggested vehicle doesn't meet the desired drop-off time is about [10\%].''

\section{Evaluation}
We evaluate the effectiveness of the explanations generated by our explainable framework in the context of five distinct vehicle routing scenarios\footnote{The dataset was obtained from Wilbur et al.~\cite{wilbur2022online}.} through an IRB-approved user study. We aim to assess the overall quality and user-friendliness of the natural language explanations from the perspective of end users, comparing them with two other baseline methods. 
Additionally, the study aims to understand the participants' preferences among three types of queries. 

\subsection{Study Design}\paragraph{Testing Environment. }Participants in this questionnaire-based user study first receive a comprehensive overview of the paratransit problem, which is designed to familiarize them with the key concepts. Then, participants are presented with five different scenarios of the route-planning problem. Each scenario involves different conditions and variables relevant to the problem, simulating real-world traffic planning situations. In the next step, the decision computed by the MCTS planning algorithm is shown to the participants. Finally, participants review three types of MCTS explanations regarding this decision. For each type of explanation, participants are asked to rate their quality using a pre-defined questionnaire. 
For all questions that require a rating, participants will be prompted to give their assessment using a 5-point Likert scale, where a rating of 1 is \textit{strongly disagree} and a rating of 5 is \textit{strongly agree}.
Illustrative examples of the questionnaire is shown in Appendix A2.5. 

\begin{algorithm}[t] 
\small
    \caption{\small \textit{ExpGen}: Generate Explanations for One Query}
    \label{alg:expgen}
    \begin{flushleft}
    \begin{algorithmic}[1] 
    
    \STATE Obtain $\phi$
    \STATE Initialize $\varepsilon$ using explanation templates
    \STATE Initialize potential violations $vio\leftarrow\{ \}$  
    \FOR{each child node $n=(s, N, Q)$ in MCTS tree} 
        \IF{$n \nvDash \phi$}
            \STATE Add $\phi$ to potential violations $vio$
        \ENDIF
    \ENDFOR
    \IF{violations $vio$ are not empty}
        \STATE Calculate explanations for $vio$ and apply to templated $\varepsilon$
    \ENDIF
    \RETURN $\varepsilon$
    \end{algorithmic}
    \end{flushleft}
\end{algorithm}

\begin{table*}[t]
\caption{Performance of the explanation framework for different query types across five scenarios.} 
\renewcommand{\arraystretch}{1.2}
\begin{adjustbox}{max width=\textwidth}
\begin{tabular}{c|ccc|ccc|ccc|ccc|ccc|ccc}
\Xhline{2pt}
\multirow{3}{*}{\textbf{Query Type}}
& \multicolumn{3}{c|}{\textbf{All Scenarios}}
& \multicolumn{3}{c|}{\textbf{Scenario 1}}
& \multicolumn{3}{c|}{\textbf{Scenario 2}}                                  
& \multicolumn{3}{c|}{\textbf{Scenario 3}}                                  
& \multicolumn{3}{c|}{\textbf{Scenario 4}}                                  
& \multicolumn{3}{c}{\textbf{Scenario 5}} \\ \cline{2-19} 
& \textbf{Det.} & \textbf{Irr.} & \textbf{Acc.}
& \textbf{Det.} & \textbf{Irr.} & \textbf{Acc.}
& \textbf{Det.} & \textbf{Irr.} & \textbf{Acc.} 
& \textbf{Det.} & \textbf{Irr.} & \textbf{Acc.} 
& \textbf{Det.} & \textbf{Irr.} & \textbf{Acc.} 
& \textbf{Det.} & \textbf{Irr.} & \textbf{Acc.} \\ \hline
\begin{tabular}[c]{@{}c@{}}Factual \\ Queries\end{tabular} 
& 4.33 & \textbf{2.40} & 3.96
& 4.39 & \textbf{2.39} & 3.99 
& 4.29 & \textbf{2.62} & 4.02 
& 4.07 & \textbf{2.11} & 3.43 
& \textbf{4.28} & \textbf{2.12} & \textbf{4.03} 
& 4.13 & \textbf{2.52} & 3.66 \\ \hline
\begin{tabular}[c]{@{}c@{}}Contrastive \\ Queries\end{tabular} 
& 4.28 & 2.50 & 3.98
& \textbf{4.43} & 2.42 & 4.11 
& \textbf{4.40} & 2.79 & 3.96 
& \textbf{4.54} & 2.36 & \textbf{4.26} 
& 4.14 & 2.41 & 3.92
& 4.35 & 2.75 & 4.01 \\ \hline
\begin{tabular}[c]{@{}c@{}}Tree \\ Expansions\end{tabular} 
& \textbf{4.35} & 2.64 & \textbf{4.08}
& 4.29 & 2.65 & \textbf{4.17}
& 4.15 & 2.69 & \textbf{4.06} 
& 3.95 & 2.33 & 3.69 
& 4.11 & 2.61 & 3.95 
& \textbf{4.39} & 2.74 & \textbf{4.13} \\ 
\Xhline{2pt}
\end{tabular}
\end{adjustbox}
\label{tab:result-by-q-type}
\end{table*}

\paragraph{Participants.} The complete details and statistics of the participants were provided in Appendix A2.2. We recruited a total number of 82 eligible participants. 
Among the participating clients, 
40.3\% of the participants reported having no knowledge of the MCTS algorithm, while 32.8\% indicated they have a basic understanding. The remaining 26.9\% of participants are either very familiar with the algorithm or have hands-on experience with MCTS. 

\subsection{Explanations Quality Assessment} 
\paragraph{Evaluation Metrics.}We designed the questionnaire following the explainable artificial intelligence metrics proposed by Hoffman et al.~\cite{hoffman2018metrics}. Specifically, we evaluate the following four \textit{criteria}: understandability, satisfaction level, completeness, and reliability employing the following survey questions: 
\textbf{(1)~Understandability:} I understand this explanation of the planning algorithm result. 
\textbf{(2)~Satisfaction:} This explanation of the planning result is satisfying.
\textbf{(3)~Completeness:} This explanation of the planning result seems complete.
\textbf{(4)~Reliability:} This explanation helps me to assess the reliability of the planning algorithm.

\paragraph{Baselines.} Our method is benchmarked against two baselines. The first baseline is designed to represent the information typically displayed in current route dispatching interfaces, while the second focuses on the states, actions, and scores stored within the MCTS search tree. \textbf{Baseline (1):} A map visualization that integrates passenger and vehicle locations, with indications of rule violations. \textbf{Baseline (2):} A detailed visualization depicting the states, actions, and scores within the search tree. 

\begin{figure}[t!]
\scriptsize
    \centering
    \begin{minipage}[b]{0.48\columnwidth}
        \centering
        \includegraphics[width=\linewidth]{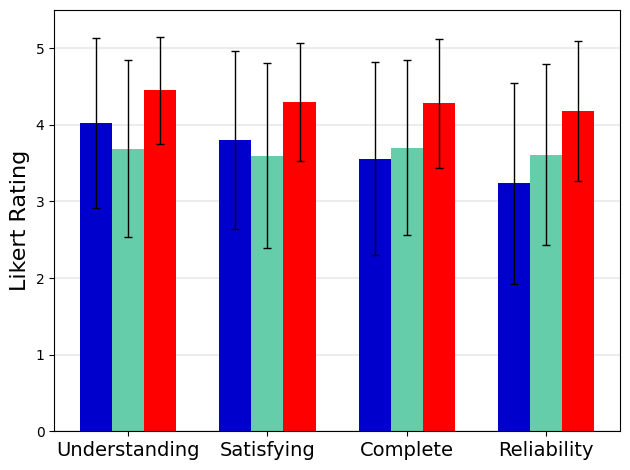}
        \textbf{(a) All Scenarios.}
        \label{fig:sub1}
    \end{minipage}
    \hfill
    \begin{minipage}[b]{0.48\columnwidth}
        \centering
        \includegraphics[width=\linewidth]{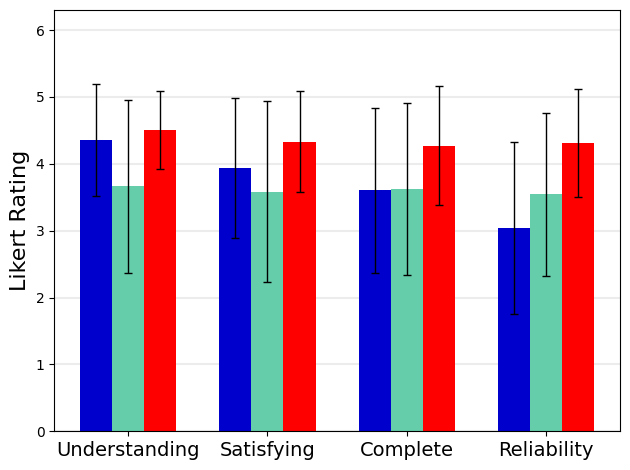}
        (b) S1: delayed arrivals.
        \label{fig:sub2}
    \end{minipage}
    \vspace{1em}
    \begin{minipage}[b]{0.48\columnwidth}
        \centering
        \includegraphics[width=\linewidth]{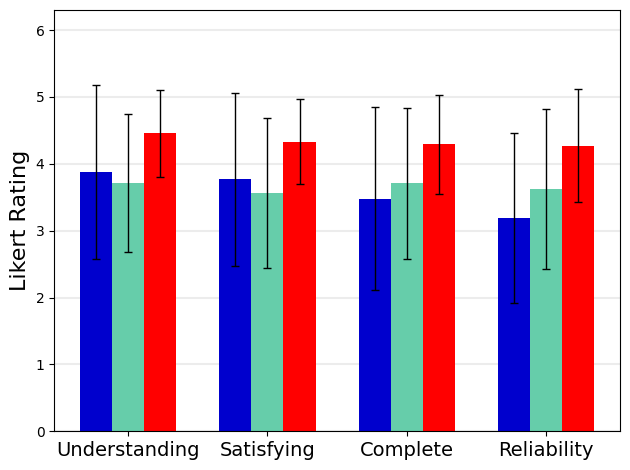}
        (c) S2: early arrivals.
        \label{fig:sub3}
    \end{minipage}
    \hfill
    \begin{minipage}[b]{0.48\columnwidth}
        \centering
        \includegraphics[width=\linewidth]{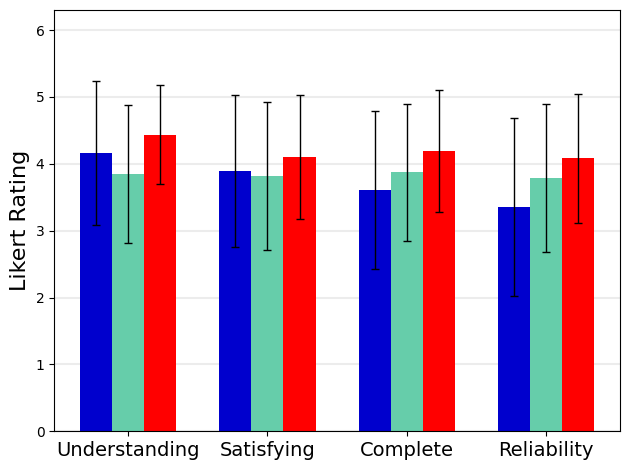}
        (d) S3: alternative plan violations.
        \label{fig:sub4}
    \end{minipage}
    \vspace{1em}
    \begin{minipage}[b]{0.48\columnwidth}
        \centering
        \includegraphics[width=\linewidth]{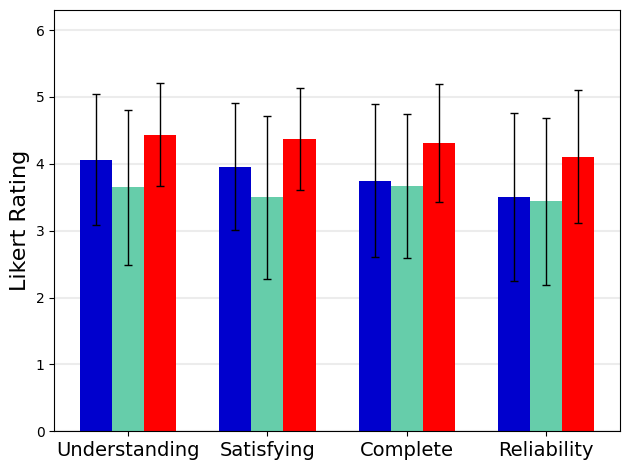}
        (e) S4: additional vehicles.
        \label{fig:sub5}
    \end{minipage}
    \hfill
    \begin{minipage}[b]{0.48\columnwidth}
        \centering
        \includegraphics[width=\linewidth]{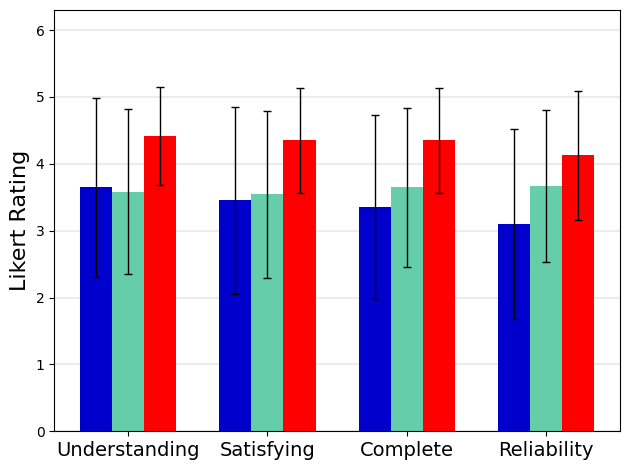}
        (f) S5: prioritizing distant vehicles.
        \label{fig:sub6}
    \end{minipage}

    \caption{Comparative analysis across five scenarios (S1-S5). \textcolor{blue}{Blue bars} represent \textit{baseline 1 with map visualization}; \textcolor{aquamarine}{green bars} represent \textit{baseline 2 with search tree visualization}; and \textcolor{red}{red bars} represent \textit{the proposed approach}. Plot (a) displays the aggregated results across all scenarios.} 
    \label{fig:five_subfigures}
  \vspace{3em}
\end{figure}

\paragraph{Assessment on explanation quality by query type.} 
We assess the effectiveness of each query type using the following questions~\cite{hoffman2018metrics}: \textbf{(1) Detail:} This explanation of the planning result has sufficient detail.
\textbf{(2) Irrelevance:} This explanation of the planning result contains irrelevant details. 
\textbf{(3) Accuracy:} This explanation says how accurate the planning algorithm is. 
The second assessment question focuses on the irrelevance of the information presented in the explanations.
Therefore, in our evaluation metrics, for the categories of detail and accuracy, a \textit{higher} score correlates with a better explanation. In contrast, for the category of irrelevance, a \textit{lower} score is more desirable, suggesting that the explanations are concise.

\begin{figure}[t]
    \centering
    \begin{minipage}{.24\textwidth}
        \centering
        \includegraphics[width=\linewidth]{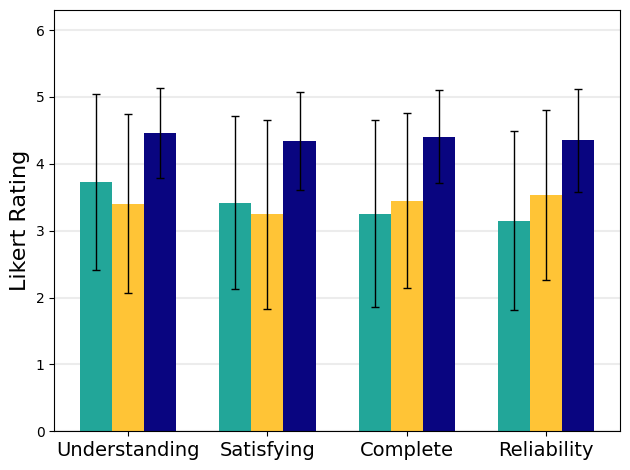}
        (a) Non-technical users.
        \label{fig:non-mcts}
    \end{minipage}%
    \hfill
    \begin{minipage}{.24\textwidth}
        \centering
        \includegraphics[width=\linewidth]{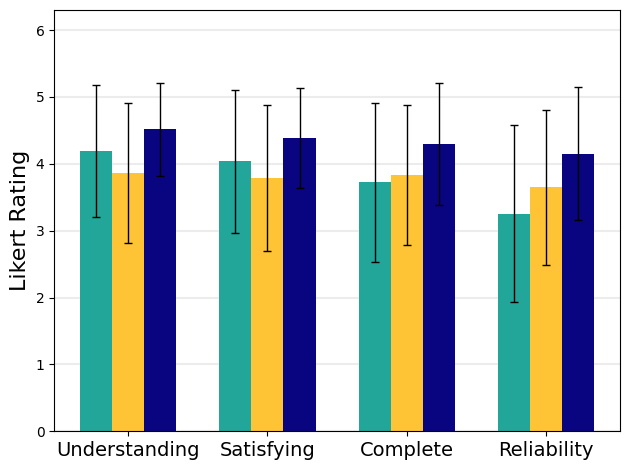}
        (b) Technical users.
        \label{fig:mcts-sub2}
    \end{minipage}
    \caption{User preferences based on technical background}
    \label{fig:user-prefs}
  \vspace{3em}
\end{figure}

\paragraph{User preferences by technical background.}
Route dispatchers, despite their knowledge in the paratransit domain, generally lack technical expertise in MCTS. Consequently, to assess the effectiveness of our explanations for both non-technical and technical users, we divided the participant population based on their technical background and familiarity with MCTS. Individuals with no knowledge about the algorithm are considered non-technical users; the rest are technical users. In the left plot of Figure~\ref{fig:user-prefs}, we present the user ratings of three baselines as evaluated by non-technical users, where green bars represent baseline 1, yellow bars represent baseline 2, and deep blue bars represent our explanations. In contrast, the right plot of Figure~\ref{fig:user-prefs} showcases the feedback from technical users.

\subsection{Result Analysis}
\paragraph{Comparison of CTL-based explainable MCTS with baselines.} In Figure~\ref{fig:five_subfigures}, we present a comparative evaluation and analysis across all scenarios. Particularly, subplot (a) aggregates and averages the Likert ratings for each criterion across the five scenarios. This result shows that, on average, our explanations outperformed both baselines in all four evaluation criteria. While participants considered baseline 1 with map visualization as more understandable and satisfying than baseline 2 with search tree visualization, the results show that the natural language-based explanations of our approach were superior in terms of understandability, satisfaction, and completeness. Furthermore, these explanations were most effective in helping participants to assess the reliability of the route planning algorithm. 
Although approximately 60\% of the participants have a basic or higher level of understanding of the MCTS algorithm, our proposed approach significantly outperformed the search tree visualization baseline in all four evaluation criteria. This performance improvement can be observed even with a technically informed participant group. 
Subplots (b)-(f) in Figure~\ref{fig:five_subfigures} showcase the results for each individual scenario. Our proposed method outperforms the other two baseline approaches across all evaluation criteria in all scenarios. Among these criteria, the most significant improvement offered by our method is in aiding participants to evaluate the reliability of the MCTS route planning algorithm. This improvement is likely attributable to the comprehensive level of detail provided by our method, as further shown by the subsequent evaluations focusing on each type of query. 


\paragraph{Evaluation of each query type.} 

In Table~\ref{tab:result-by-q-type}, we present the average Likert ratings for three key evaluation criteria across all five scenarios. Note that in assessing irrelevance, a lower rating is more favorable. 
Across all five scenarios, the result reveals that contrastive queries are considered as having the most sufficient detail, particularly in explaining why certain requests were not assigned to a vehicle. 
Moreover, participants rated the factual queries as the most relevant, highlighting their ability to convey essential information without including unnecessary details. 
About the evaluation of the accuracy of the MCTS algorithm, participants showed a preference for queries that require tree expansion. This preference suggests that they find the information through additional search, beyond what is available in the original search tree, is helpful in further understanding the accuracy of MCTS. 

\paragraph{User preferences by technical background.} Data in these figures reveal that our explainer received higher ratings across all four evaluation criteria from both user groups. Notably, the advantage of our explainer is more pronounced in the feedback from non-technical users, indicating its effectiveness and adaptability in meeting the needs of users with little to no technical expertise. 

\subsection{Discussion}
\paragraph{Learning curve. } 
In Figure~\ref{fig:five_subfigures}, we observe a trend where the performance gap between our proposed method and the two baseline methods widens progressively from scenario 1 to scenario 5. This trend could be attributed to participants experiencing a learning curve in using our system and gaining increased familiarity with the route planning task. 
Initially, the map visualization might seem adequate for understanding why a particular request is assigned to a vehicle, especially for non-technical users. However, as participants look deeper into the more detailed natural language explanations provided by our explainer, they begin to understand more about the complexity of the scenarios, which is often more complicated than initially perceived. 
The increasing complexity of the scenarios, especially in the last two tasks where the number of paratransit vehicles grows, likely further emphasizes the superiority of our proposed method over the map visualization. 
Interestingly, we observe that participants' comprehension of the search tree visualization does not show marked improvement with increased exposure. This insight underscores the need for explanations that can adapt to the complexity of the scenario and the user's growing understanding. 

\paragraph{Additional feedback. } To deepen our understanding of user preferences regarding the explanations provided, we included additional open-ended questions at the end of the survey. 
For instance, one participant positively highlighted our proposed system, stating that it \textit{``provides me more useful information on how to interpret what the algorithm is doing.''} This feedback emphasizes the effectiveness of our system in enhancing user comprehension of the algorithm's processes. 
Several participants highlighted the effectiveness of our explainer in clarifying the routing decisions in scenario 5, particularly why the closer vehicle with ID 1 was not chosen, while a more distant vehicle with ID 3 was assigned instead. 
For instance, one participant remarked, ``I like the solution that was given this time; it clearly explains that vehicle 3 should be taken over vehicle 1.'' These comments underscore the explainer's ability to provide reasons behind complex decision-making processes in a comprehensible way. 
\section{Related Work}

Explainable AI research has its roots in the need to comprehend, trust, and enhance AI algorithms~\cite{arrieta2020explainable}. Recent studies have highlighted the importance of assisting human users in comprehending and trusting the decisions made by a wide range of models, including regression trees, classification models, and neural networks (e.g.~\cite{das2020opportunities,samek2019towards,georgiev2022algorithmic,angelov2021explainable}).
In general, our approach is situated within the broader context of prior research centered on explaining plans and decisions generated by computer algorithms. While there is currently no comprehensive taxonomy for this research domain, we can broadly categorize these efforts as follows. 

The first category is related to result-oriented explanations, where the objective of explainability algorithms is to reveal the rationale behind a specific decision post hoc by considering the attributes of that decision. Previous work, such as Langley~\cite{langley2016explainable}, discusses the desired behaviors of post-hoc explainability modules at a high level, encompassing features like explaining the objectives of the planning task and presenting alternative plans. 
A comprehensive survey, Sreedharan et al.~\cite{sreedharan2020emerging}, formally defines distinct primary considerations of explainability systems. These considerations are roughly categorized based on the intended audience (the ``explainee'') and the nature of the planning results, which can encompass plans or policies.
Furthermore, Fox et al.~\cite{fox2017explainable} identify four key types of questions that should be addressed by an explanation system. They illustrate the practical application of the system with two examples, demonstrating how it should function in practice. 
Nonetheless, the majority of these papers discuss post-hoc explainability systems in a general manner. They often lack specific examples and use cases that showcase explanations in action. Thus, the development of meaningful solutions motivated by real-world problems remains a significant challenge.
In contrast, our work is the first to develop an explanation framework for a public transportation-related planning system in the real world, and evaluates the system with human users. 

The second category aims to generate explanations that are included as a sub-goal of the planning algorithm. For instance, the explainability feature may influence the planning process, where the objective is to also account for system interactions with other system components~\cite{sreedharan2020bridging}. Chakraborti~\cite{chakraborti2017plan} views the planning problem as a system with two components: the first component is the AI algorithm, and the second component is the human agent. The goal of the AI algorithm is not to make decisions but to provide suggestions to the human. 
Similar approaches can also be found in the field of multi-agent reinforcement learning (MARL). Boggess et al.~\cite{boggess2022toward} focus on explaining MARL policies by providing policy summaries in natural language. To ensure comprehensive explanations for user queries, their system conducts additional guided rollouts of the policy to generate informative explanations. Our work shares similar practices in providing explanations for under-explored tree branches with the help of additional computation. However, our approach not only generates more fine-grained explanations leveraging CTL and quantitative evaluations, but also derives explanations from the search tree of an online planner as opposed to a fixed RL policy. 
\section{Conclusions}

In this study, we present a CTL-based explainer specifically designed for the MCTS algorithm, focusing on the paratransit route planning application. Our explainer classifies user queries into three categories: factual queries, contrastive queries, and queries requiring path explanations. It then dynamically generates CTL formulas based on the free variables, the type of query and its specifications. After checking different branches of the existing MCTS search tree or even the results of newly initiated searches, the explainer converts the outcomes of its specification verification into natural language explanations with the help of language templates. This approach is designed to enhance the usability of the system for both technical and non-technical audiences. The effectiveness of our method is evidenced by a user study involving 82 participants, which shows notable improvements in user comprehension and satisfaction compared to two baseline methods. We believe that our explainer is versatile enough to be seamlessly adapted to other MCTS implementations in various application fields, broadening its potential for impact.

\section*{Acknowledgments}
This material is based upon work supported by the National Science Foundation (NSF) under Award Numbers 2028001, 2220401, CNS-2238815 and CNS-1952011, AFOSR under FA9550-23-1-0135, and DARPA under FA8750-23-C-0518. The authors acknowledge the support and guidance of Philip Pugliese from Chattanooga Area Regional Transportation Authority. This research has received funding from the project ALIGN4Energy (NWA.1389.20.251) of the research programme NWA ORC 2020 which is (partly) financed by the Dutch Research Council (NWO), and from the European Union’s Horizon Europe Research and Innovation Programme, under Grant Agreement number 101120406. The paper reflects only the authors’ view and the EC is not responsible for any use that may be made of the information it contains.

\bibliography{aaai24}

\newpage
\section*{Technical Appendix}

In this technical appendix, we include supplementary information about our approach and details of the survey experiment. Section A1 expands upon Section 3 of the main text, presenting further details such as the conversion of queries to CTL and a pseudocode. In Section A2, we present our survey design, interface, query and answer templates utilized in the questionnaire, and additional quantitative results.

\section*{A1. Additional Details on Section 3}
In Table~\ref{tab:type_to_query1} and Table~\ref{tab:query_to_ctl2}, we present additional examples demonstrating how queries related to transit planning tasks can be formulated and translated into CTL formulas, which supplement the information provided in Table 1 of the main text. In Algorithm 3, we provide the pseudocode for generating additional search processes for queries that requires tree expansion. 
\begin{algorithm}[H]
\small
\caption{\small \textit{TreeExp}: Tree expansion for additional search.}
\label{alg:expandmcts}
\begin{algorithmic}[1]
\REQUIRE original MCTS tree $T$, queried node $N_q$
\STATE Obtain $T.RootNode$ 
\STATE Define node to be expanded $N_e \leftarrow T.$enumerate($N_q$) 
\STATE Initialize new tree $T_n$ 
\STATE Let current node $\leftarrow$ $N_e$
\STATE $T_n$.addNode($N_e.state$, $N_e.action$)
\STATE Run MCTS($N_e$)
\RETURN new MCTS tree $T_n$
\end{algorithmic}
\end{algorithm}

\section*{A2. Additional Details on Survey Design}
\subsection*{A2.1. Study Procedures} 
Our study included the following procedures: 
\begin{itemize}
    \item Participants would review a description of the paratransit problem, which is a specialization of the vehicle routing problem and includes serving trip requests from passengers.
    \item Participants were given: several scenarios, one after the other; the decision computed by the planning algorithm; multiple types of explanations as to ``why'' the algorithm chose the decision. One of the types of explanations is interactive, i.e., they could query the algorithm through a set of pre-defined questions.
    \item Participants' goal was to evaluate which form of explanation helps them the most. We asked them to rate the explanations through a pre-defined questionnaire.
\end{itemize}

\begin{table}[t]
    \centering
    \caption{Examples of queries with specification types by query type.}
    \begin{tabular}{L{1.95cm}|L{3.1cm}|L{2.05cm}}
    \toprule
      \textbf{Query Type} & \textbf{Query Example} & \textbf{ Specification Type} \\
      \midrule
      \textbf{T1:} Factual queries & \textbf{Q1:} Is it expected that the [passenger] will be [picked up] on time? & Efficiency \\
      \midrule 
      \textbf{T1:} Factual queries & \textbf{Q2:} Is the route compliant with vehicle [capacity] constraints? & Hard constraint \\
      \midrule 
      \textbf{T1:} Factual queries & \textbf{Q3:} Does the vehicle have sufficient [fuel] to complete the route? & Hard constraint \\
      \midrule 
      \textbf{T2:} Contrastive queries & \textbf{Q4:} Why wasn't the passenger assigned to [this alternative vehicle], which is closer to the passenger? & Hard Constraint; \newline Efficiency \\
      \midrule
      \textbf{T3:} Queries with path expansion & \textbf{Q5:} Can you tell me more about [this alternative route]? & Hard Constraint; \newline Efficiency; Soundness \\
      \midrule
      \textbf{T3:} Queries with path expansion & \textbf{Q6:} If you think more about [this other option], would you change your recommendation? & Hard Constraint; \newline Efficiency; Soundness \\
      \bottomrule
    \end{tabular}
    \label{tab:type_to_query1}
\end{table}

\begin{table}[t]
    \centering
    \caption{Example specification types with CTL formulas}
    \begin{tabular}{L{2cm}|L{3.1cm}|L{2cm}}
    \toprule
      \textbf{Specification Type} &  \textbf{State Variable} &  \textbf{CTL Formula} \\
    \midrule
        Efficiency & $t_{\text{est}}$ estimated travel time;\newline $t_p$ request picked up time;\newline $t_a$ allowed time window & $\phi_1: AG \, ( t_{\text{est}} \leq (t_p + t_a))$ \\
      \midrule 
      Efficiency & $t_{\text{est}}$ estimated travel time;\newline $t_d$ request drop off time;\newline $t_a$ allowed time window  & $\phi_2: AG \, ( t_{\text{est}} \leq (t_d + t_a))$ \\
      \midrule Hard Constraint & $v_c$ vehicle capacity;\newline $v_o$ vehicle occupancy & $\phi_3: AG \, (v_o \leq v_c )$ \\
      \midrule
      Hard Constraint & ${v}_{\text{ft}}$ fuel tank reading;\newline ${v}_{\text{fr}}$ fuel required & $\phi_4: AG \, ({v}_{\text{fr}} \leq {v}_{\text{ft}})$\\
      \midrule
      Soundness & ${v}_{\text{tt}}$ total travel time;\newline ${v}_{\text{rt}}$ reasonable timeframe & $\phi_5: AG \, ({v}_{\text{tt}} \leq {v}_{\text{rt}})$\\
      \midrule
      Soundness & ${\theta}_{s}$ intermediate stops;\newline ${\theta}_{d}$ reasonable number of stops & $\phi_6: AG \, ({\theta}_{s} \leq {\theta}_{d}$\\
      \midrule
      Soundness & ${r}_{\text{cs}}$ request current status; & $\phi_7: AF \, ({r}_{\text{cs}} = \text{dropped-off})$\\
      \bottomrule
    \end{tabular}
    \label{tab:query_to_ctl2}
\end{table}

\subsection*{A2.2. Participants} We recruited a total number of 82 eligible participants. We identify and contact potential participants via a mass email network and institute Slack messages. 
Among the participating clients, 36.9\% identified as women, 58.5\% as men, and 4.6\% preferred not to disclose their gender. Additionally, 73.1\% of the participants have a basic understanding of public transportation but lack familiarity with the para-transit domain, while 26.9\% are knowledgeable about or have experience with para-transit and vehicle routing. Regarding their technical expertise with the MCTS algorithm, 40.3\% of the participants reported having no knowledge of the algorithm, while 32.8\% indicated they have a basic understanding. The remaining 26.9\% of participants are either very familiar with the algorithm or have hands-on experience with MCTS. 

\subsection*{A2.3. Survey Scenario Descriptions}
In our official online survey, participants received the following information for each scenario: 

\textit{This map displays the requested pickup location (shown by a human figure) and the current vehicle positions (shown by bus/shuttle icons). As a route dispatcher, you manage vehicles that are shared by multiple passengers at the same time. The recommended routing decision is: The computer algorithm assigned the passenger to the vehicle, which is indicated by an arrow pointing to the passenger; i.e., the vehicle that has an arrow to the passenger will pick up the passenger and drop off the passenger.} 

\textit{In the following maps, you will see a simplified paratransit dispatch system. These visualizations show key components, including the positions of vehicles and passengers, recommendations generated by the assignment algorithm, and constraint violation (if any). }

\begin{figure}[h]
\centering
  \includegraphics[width=0.9\linewidth]{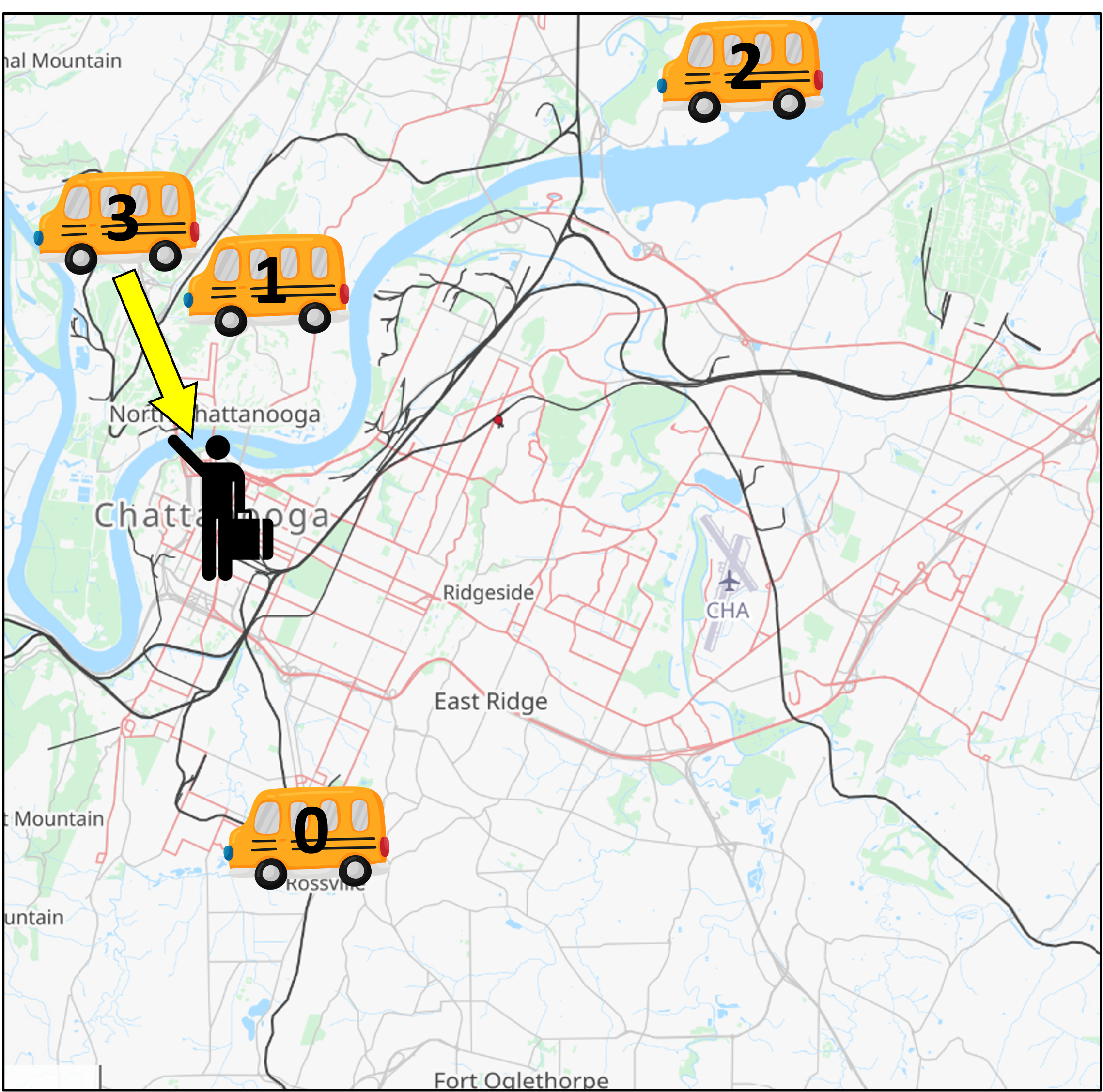}
  \caption{An example of the survey scenarios. }
  \label{fig:surveyexample}
  \vspace{3em}
\end{figure}

\subsection*{A2.4. Survey Questions}
\paragraph{Pre-Survey Questions.}
\begin{itemize}
    \item How familiar are you with the paratransit domain?
    \begin{enumerate}[label=\alph*)]
        \item \textit{I have heard about public transportation, but I do not know about vehicle routing problems or paratransit.}
        \item \textit{I was already familiar with the rules of the paratransit domain.}
        \item \textit{I have solved other vehicle routing problems, but I am not familiar with paratransit.}
    \end{enumerate}
    
    \item How familiar are you with Monte Carlo Tree Search algorithm?
    \begin{enumerate}[label=\alph*)]
        \item \textit{I do not know what Monte Carlo tree search is.}
        \item \textit{I have a basic knowledge of the algorithm, but I do not know the details.}
        \item \textit{I am well aware of how the algorithm works.}
        \item \textit{I have practical experience with Monte Carlo tree search; I've used or coded it at least once.}
    \end{enumerate}
    
    \item To which gender identity do you most identify? 
    \textit{Note:} This question is for demographic purposes only. Your response to this question will not affect our interpretation of your feedback.
    \begin{enumerate}[label=\alph*)]
        \item \textit{Female}
        \item \textit{Male} 
        \item \textit{Transgender}
        \item \textit{Non-binary}
        \item \textit{Prefer not to answer} 
        \item \textit{Other} 
    \end{enumerate}
\end{itemize}

\paragraph{Survey Question Set 1: Comparing with Baselines.}

\begin{itemize}
    \item I understand this explanation of the planning result. 
    \item This explanation of the planning result is satisfying. 
    \item This explanation of the planning result seems complete. 
    \item This explanation helps me to assess the reliability of the planning algorithm. 
\end{itemize}

\paragraph{Survey Question Set 2: Evaluating Query by Types.}
\begin{itemize}
    \item This explanation of the planning result has sufficient detail. 
    \item This explanation of the planning result contains irrelevant details. 
    \item This explanation says how accurate the planning algorithm is. 
\end{itemize}

\paragraph{Survey Question Set 3: Open-Ended Questions.}
\begin{itemize}
    \item If given the chance, would you like to ask more questions? If so, please specify.
    \item Is there anything about the descriptive text explanations shown that you would suggest changing? If so, please specify. 
\end{itemize}

\paragraph{Survey Question Choice.}
\begin{enumerate}[label=\alph*)]
    \item \textit{Strongly agree.} 
    \item \textit{Agree.} 
    \item \textit{Neutral.} 
    \item \textit{Disagree.} 
    \item \textit{Strongly disagree.} 
\end{enumerate}

\subsection*{A2.5. Example User Interface}
\begin{figure}[t]
\centering
  \includegraphics[width=0.95\linewidth]{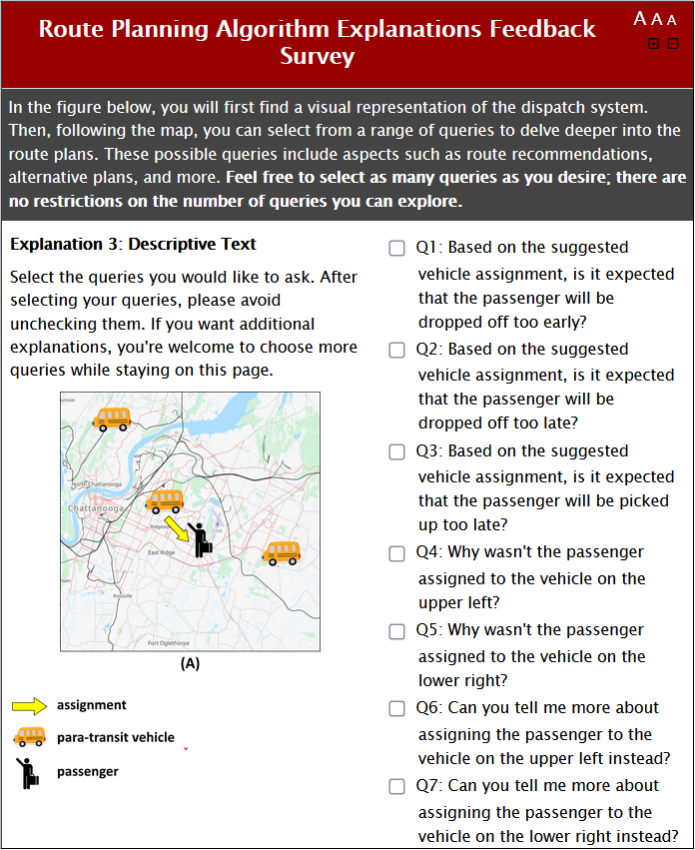}
  \caption{An example survey interface. }
  \label{fig:interf}
  \vspace{3em}
\end{figure}

Our study utilized the Research Electronic Data Capture (REDCap) platform, which is a secure web-based application designed for creating and managing online surveys. Figure~\ref{fig:interf} illustrates an example screenshot of the user interface from our online survey. 

\begin{table*}[t]
    \centering
    
    \begin{minipage}{.5\textwidth}
        \centering
        Performance for different query types across all scenarios.
        \begin{tabular}{lccc}
            \hline
            \textbf{Query Type} & \textbf{Detail} & \textbf{Irrelevance} & \textbf{Accuracy} \\ \hline
            Factual Queries & 4.33$\pm$1.00 & 2.40$\pm$1.27 & 3.96$\pm$1.13 \\
            Contrastive Queries & 4.28$\pm$0.89 & 2.50$\pm$1.30 & 3.98$\pm$1.03 \\
            Tree Expansions & 4.35$\pm$0.80 & 2.64$\pm$1.37 & 4.08$\pm$0.94 \\
            \hline
        \end{tabular}
    \end{minipage}%
    \begin{minipage}{.5\textwidth}
        \centering
        Performance for different query types for Scenario 1.
        \begin{tabular}{lccc}
            \hline
            \textbf{Query Type} & \textbf{Detail} & \textbf{Irrelevance} & \textbf{Accuracy} \\ \hline
            Factual Queries & 4.39$\pm$0.66 & 2.39$\pm$1.02 & 3.99$\pm$0.88 \\
            Contrastive Queries & 4.43$\pm$0.70 & 2.42$\pm$1.44 & 4.11$\pm$1.16 \\
            Tree Expansions & 4.29$\pm$0.83 & 2.65$\pm$1.34 & 4.17$\pm$0.92 \\
            \hline
        \end{tabular}
    \end{minipage}

    \vspace{2em}
    \begin{minipage}{.5\textwidth}
        \centering
        Performance for different query types for Scenario 2.
        \begin{tabular}{lccc}
        \hline
        \textbf{Query Type} & \textbf{Detail} & \textbf{Irrelevance} & \textbf{Accuracy} \\ \hline
        Factual Queries & 4.29$\pm$0.97 & 2.62$\pm$1.47 & 4.02$\pm$1.05 \\
        Contrastive Queries & 4.40$\pm$0.93 & 2.79$\pm$1.38 & 3.96$\pm$0.98 \\
        Tree Expansions & 4.15$\pm$0.94 & 2.69$\pm$1.48 & 4.06$\pm$1.07 \\
        \hline
        \end{tabular}
    \end{minipage}%
    \begin{minipage}{.5\textwidth}
        \centering
        Performance for different query types for Scenario 3.
        \begin{tabular}{lccc}
        \hline
        \textbf{Query Type} & \textbf{Detail} & \textbf{Irrelevance} & \textbf{Accuracy} \\ \hline
        Factual Queries & 4.07$\pm$1.20 & 2.11$\pm$1.16 & 3.43$\pm$1.35 \\
        Contrastive Queries & 4.54$\pm$0.60 & 2.36$\pm$1.36 & 4.26$\pm$0.83 \\
        Tree Expansions & 3.95$\pm$1.22 & 2.33$\pm$1.16 & 3.69$\pm$1.29 \\
        \hline
        \end{tabular}
    \end{minipage}

    \vspace{2em}
    \begin{minipage}{.5\textwidth}
        \centering
        Performance for different query types for Scenario 4.
        \begin{tabular}{lccc}
        \hline
        \textbf{Query Type} & \textbf{Detail} & \textbf{Irrelevance} & \textbf{Accuracy} \\ \hline
        Factual Queries & 4.28$\pm$1.05 & 2.12$\pm$1.31 & 4.03$\pm$1.09 \\
        Contrastive Queries & 4.14$\pm$1.00 & 2.41$\pm$1.13 & 3.92$\pm$0.95 \\
        Tree Expansions & 4.11$\pm$1.10 & 2.61$\pm$1.29 & 3.95$\pm$0.97 \\
        \hline
        \end{tabular}
    \end{minipage}%
    \begin{minipage}{.5\textwidth}
        \centering
        Performance for different query types for Scenario 5. 
        \begin{tabular}{lccc}
        \hline
        \textbf{Query Type} & \textbf{Detail} & \textbf{Irrelevance} & \textbf{Accuracy} \\ \hline
        Factual Queries & 4.13$\pm$1.19 & 2.52$\pm$1.40 & 3.66$\pm$1.32 \\
        Contrastive Queries & 4.35$\pm$0.60 & 2.75$\pm$1.35 & 4.01$\pm$0.93 \\
        Tree Expansions & 4.39$\pm$0.61 & 2.74$\pm$1.39 & 4.13$\pm$0.83 \\
        \hline
        \end{tabular}
    \end{minipage}
\end{table*}

\begin{figure}[t]
\centering
  \includegraphics[width=0.9\linewidth]{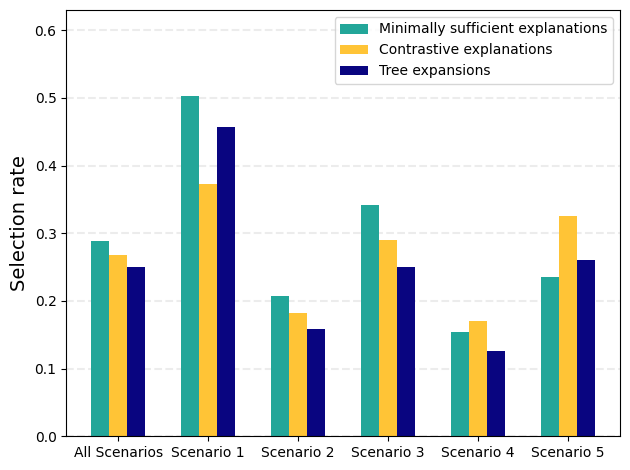}
  \caption{Query selection rate by type. }
  \label{fig:sel-rate}
  \vspace{3em}
\end{figure}

\subsection*{A2.6. Queries and Explanation Template Examples}

\begin{enumerate}
    \item \textbf{Q1:} Based on the suggested vehicle assignment, is it expected that the passenger will be [dropped off too late]? 
    \begin{itemize}
        \item The passenger has specified a desired drop-off time of [5:33]. In order to determine the most efficient route, the Monte Carlo Tree Search (MCTS) route planning algorithm simulates approximately [150] potential future scenarios and requests. This thorough exploration helps the algorithm to make an informed recommendation. 
        \item Potential Late Arrival: Based on the extensive set of scenarios examined by MCTS, there's a chance that the passenger might encounter a delay in their drop-off time. This anticipated delay averages around [23 minutes]. The primary reason for this delay is that the proposed vehicle is expected to make stops at about [4 other locations] prior to reaching the passenger's drop-off point. However, the delay can be as short as [19 minutes] or extend up to [27 minutes]. The percentage of times the suggested vehicle doesn't meet the desired drop-off time is about [10\%].
    \end{itemize}

    \item \textbf{Q2:} Why wasn't the passenger assigned to [the vehicle on the left]? 
    \begin{itemize}
        \item The MCTS route planning algorithm provides recommendations based on a simulation of various future scenarios. It assesses the potential outcomes of assigning a specific request to different vehicles by assigning them ``scores.'' When comparing this alternative vehicle to the recommended one, the latter has a composite score of [192], while the former scored at [35]. This lower score exhibited by the alternative vehicle suggests suboptimal performance, making it less favorable for the task at hand. Why is the recommended vehicle better? Here are two main reasons:
        \item More Trips: The recommended vehicle demonstrates a more than [400\%] enhancement in the service rate, meaning that it can possibly handle more trips without getting overwhelmed.
        \item On-Time Service: The recommended vehicle demonstrates a more than [450\%] enhancement in the punctuality, meaning that passengers are more likely to get to their destination right on time.
    \end{itemize}
    
    \item \textbf{Q3:} Can you tell me more about assigning the passenger to [the vehicle on the lower right] instead? 
    \begin{itemize}
        \item To answer your query, our MCTS route planning system dove deeper into its decision-making process. It examines various possible ``futures'' to provide more information about the alternative plan. 
        \item More Scenarios Analyzed: MCTS looked at [74] new future traffic and route situations that it hadn't considered before. Results from the New Analysis: Even with this deeper look, the system found that the alternate plan still wasn't the best choice overall. In the specific situation you asked about, the passenger might arrive too early, by about [33 minutes]. Consistency Check: To ensure the consistency of this information, we checked how often this early drop-off happens in all the new scenarios: this happens in [84\%] of them. 
    \end{itemize}
\end{enumerate}

\subsection*{A2.7. Additional Results}

\paragraph{Query Selection by Type.}
In Figure 9, we offer statistics on the selection rates of different query types across Scenarios 1-5, as well as an aggregation of all scenarios. The data shows that in simpler scenarios (Scenarios 1-3), factual queries are predominantly chosen by users, as highlighted by the green bars. In contrast, in more complex scenarios that involve a larger number of vehicles and require more intricate sequential decision-making, there is a significant increase in the selection of contrastive explanations. This result suggests that users find contrastive explanations increasingly beneficial as the complexity of the scenario escalates. 

\paragraph{Supplementary to Table 2.}
In the tables shown above, we present detailed statistics comparing each query type across different scenarios, supplementing the information provided in Table 2 of the main text. We also include the standard deviation for these comparisons.

\end{document}